\documentclass[letterpaper, 10 pt, conference]{ieeeconf}  

\IEEEoverridecommandlockouts                              

\usepackage{graphics} 
\usepackage{epsfig} 
\usepackage{amsmath} 
\usepackage{amssymb}  
\usepackage{algorithm}
\usepackage{verbatim}
\usepackage[noend]{algpseudocode}
\usepackage{subfig}
\let\eqref\undefined

\newcommand{\eqref}[1]{Eqn.~\ref{eq:#1}}

\newcommand{\eg}[1]{\textit{e.g.,}}
\newcommand{\etc}[1]{\textit{etc.}}
\newcommand{\vs}[1]{\textit{vs.}}

\title{\LARGE \bf Curating Long-Term Vector Maps}

\author{Samer Nashed$^{1}$ and Joydeep Biswas$^{2}$
\thanks{$^{1}$Samer Nashed and $^{2}$Joydeep Biswas are with the College of Information and Computer Sciences,
        University of Massachusetts, Amherst, MA 01003, USA.
        Email: {\tt\small \{snashed, joydeepb\}@cs.umass.edu}}}%

\begin{document}

\bibliographystyle{abbrv}

\maketitle
\thispagestyle{empty}
\pagestyle{empty}

\begin{abstract}
Autonomous service mobile robots need to consistently, accurately, and robustly
localize in human environments despite changes to such environments over time.
Episodic non-Markov Localization addresses the challenge of localization in such
changing environments by classifying observations as arising from Long-Term,
Short-Term, or Dynamic Features. However, in order to do so, EnML relies on an
estimate of the Long-Term Vector Map (LTVM) that does not change over time. In this paper, we introduce a recursive algorithm to build and update the LTVM over time by reasoning about visibility constraints of objects observed over multiple robot deployments. We use a signed distance function (SDF) to filter out observations of short-term and dynamic features from multiple deployments of the robot. The remaining long-term observations are used to build a vector map by robust local linear regression. The uncertainty in the resulting LTVM is computed via Monte Carlo resampling the observations arising from long-term features.
By combining occupancy-grid based SDF filtering of observations with continuous space regression of the filtered observations, our proposed approach builds, updates, and amends LTVMs over time, reasoning about all observations from all robot deployments in an environment. We present experimental results demonstrating the accuracy, robustness, and compact nature of the extracted LTVMs from several long-term robot datasets.
\end{abstract}

\section{Introduction}

Long-term autonomy is an essential capability for service mobile robots deployed in human environments. One challenging aspect of long-term autonomy
in a human environment is robustness to changes in the environment. 

\begin{figure}[!ht]
  \centering
  \includegraphics[scale=0.3, trim = 0mm 10mm 0mm 0mm]{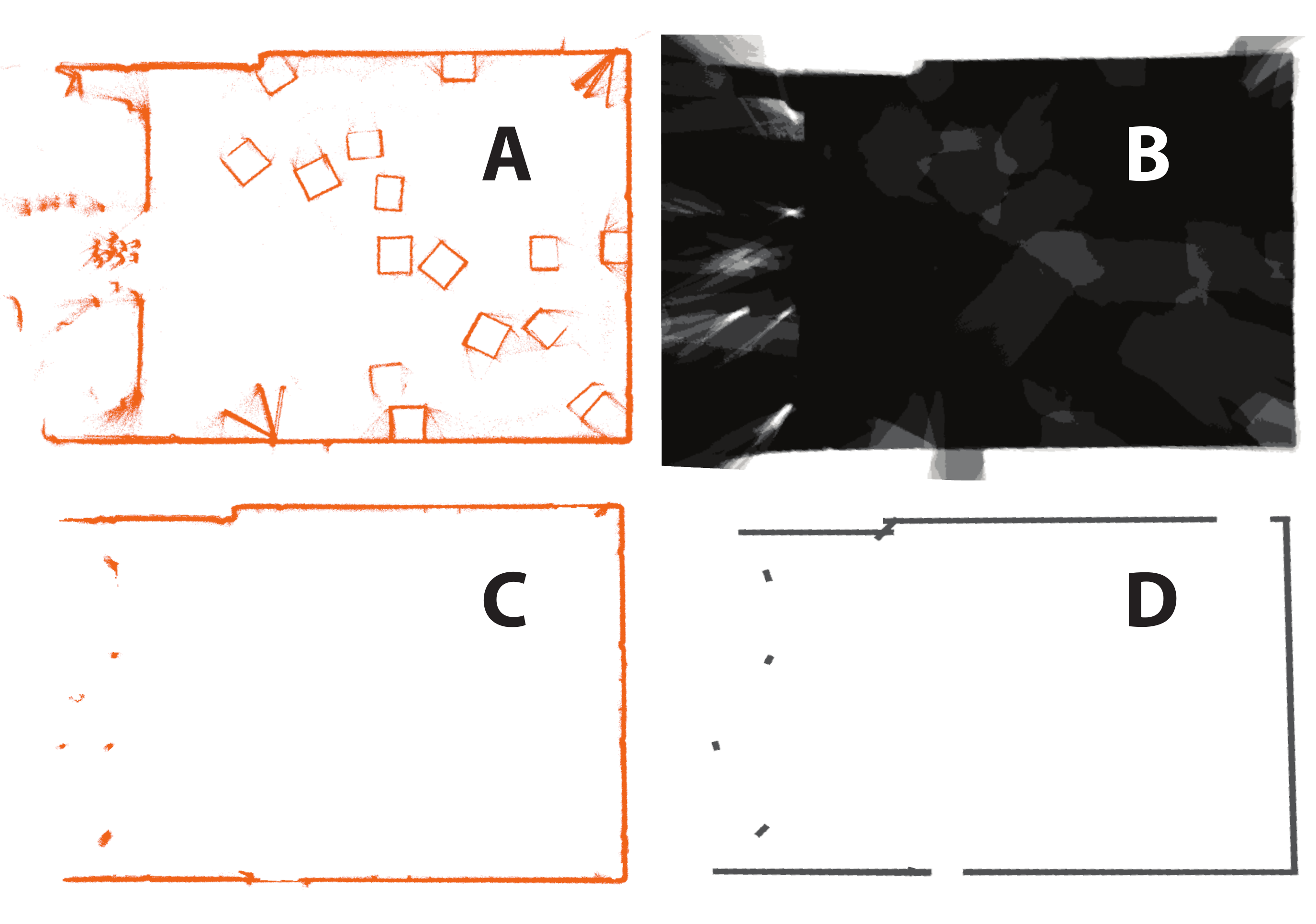}
  \caption{Observations at different stages of the LTVM pipeline. In alphabetical order: raw data from all deployments, weights computed by the SDF, filtered data, final LTVM.}
  \vspace{-2.0em}
  \label{fig:Teaser}
\end{figure}

Many approaches have been proposed to reason about a changing environment, including estimating the latest state of the environment~\cite{dpgslam, spectral}, estimating different environment configurations~\cite{recency, stach}, or modeling the dynamics of the environment~\cite{omnse, iMac, hmm, TOG, TMDE}. However, in large environments, it may not be feasible to make the periodic observations required by these approaches. Therefore, we model observations in human environments as arising from three distinct types of features~\cite{EnML}: Dynamic Features (DFs) or moving objects such as people or carts, Short- Term Features (STFs) or movable objects such as tables or chairs, and Long-Term Features (LTFs) which persist over long periods of time, such as office walls or columns. Episodic non-Markov Localization (EnML)~\cite{EnML} simultaneously reasons about global localization information from LTFs, and local relative information from STFs. A key requirement to EnML is an estimate of the Long-Term Vector Map: the features in the environment that persist over time, represented in line segment or vector form (Fig. \ref{fig:Teaser}). 


In this paper, we introduce an algorithm to build and update Long-Term Vector Maps indefinitely, utilizing observations from all deployments of all the robots in an environment. Our proposed algorithm  filters out observations corresponding to DFs from a single deployment using a signed distance function (SDF)~\cite{SDF}. Merging the SDFs from multiple deployments then filters out the short-term features. Remaining observations correspond to LTFs, and are used to build a vector map via robust local linear regression. Uncertainty estimates of the resultant Long-Term Vector Map are calculated by a novel Monte Carlo uncertainty estimator. Our proposed approach thus consists of the following steps:
\begin{enumerate}
  \item \emph{Filter:}  Use the most recent observations to compute an SDF and discard points based on weights and values given by the SDF (Section IV).
  \item \emph{Line Extraction:} Use greedy sequential local RANSAC \cite{RANSAC} and non-linear least-squares fitting to extract line segments from the filtered observations (Section V).
  \item \emph{Estimate Feature Uncertainty:} Compute segment endpoint covariance estimates via Monte Carlo resampling of the observations (Section VI).
  \item \emph{Map Update:} Add, merge, and delete lines using a de-coupled scatter matrix representation~\cite{scatter} (Section VII).
\end{enumerate}

Our approach takes advantage of the robust filtering provided by the SDF while avoiding dependency on a discrete world representation and grid resolution by representing LTFs as line segments in $\mathbb{R}^2$. Evaluation of our approach is detailed in section VIII. Across all metrics examined in this study, we find vector maps constructed via SDF filtering comparable or favorable to occupancy grid based approaches.

\section{Related Work}

The problem of long-term robotic mapping has been studied extensively, with most algorithms relying on one of two dominant representations: occupancy grids~\cite{og1, og2} and geometric or polygonal maps~\cite{opm, lmap}. Recently, work towards metric map construction algorithms that are able to cope with dynamic and short-term features has accelerated. 

Most approaches fall into one of four categories: dynamics on occupancy grids, latest state estimation, ensemble state estimation, and observation filters.

One common approach models map dynamics on an occupancy grid using techniques such as learning non-stationary object models~\cite{omnse} or modeling grid cells as Markov chains~\cite{iMac, hmm}. Alternatively, motivated by the widely varying timescales at which certain aspects of an environment may change, some approaches seek to leverage these differences by maintaining information relating to multiple timescales within one or more occupancy grids~\cite{TOG, TMDE}. 

Other approaches estimate the latest state of the world, including dynamic and short-term features. Dynamic pose-graph SLAM~\cite{dpgslam} can be used in low-dynamic environments, and spectral analysis techniques~\cite{spectral} attempt to predict future environment states on arbitrary timescales. 

Instead of estimating solely the latest state, some approaches estimate environment configurations based on an ensemble of recent states. Temporal methods such as recency weighted averaging~\cite{recency} determine what past information is still relevant, and other techniques such as learning local grid map configurations~\cite{stach} borrow more heavily from the dynamic occupancy grid approach.

Another approach filters out all observations corresponding to non-LTFs, resulting in a ``blueprint" map. Previous algorithms have had some success filtering dynamic objects, specifically people~\cite{PeopleFilter}, but have struggled to differentiate between STFs and LTFs. Furthermore, all of the methods mentioned above rely on an occupancy grid map representation, whereas our method produces a polygonal, vector map.

\section{Long-Term Vector Mapping}

Long-term vector mapping runs iteratively over multiple robot deployments, operating on the union of all registered laser observations from the given deployment, aligned to the same frame. We call these unions composite scans and denote them $C = \cup_{i=1} ^N S_i$, where $S_i$ is a single laser scan. Composite scans are processed in batch after each deployment, and may be generated via Episodic non-Markov Localization ~\cite{EnML} or a similar localization algorithm. 

After each deployment, a short-term signed distance function (ST-SDF) given by a set of weights $W'$ and values $V'$ is computed over the area explored by the robot by considering the composite scan $C$. The ST-SDF is then used to update the long-term SDF (LT-SDF), given by $W$ and $V$, which aggregates information over all previous deployments. The updated LT-SDF is denoted $W^*$ and $V^*$, and is used to determine a filtered composite scan $C' \subset C$, containing observations corresponding exclusivley to LTFs. We call this process SDF-filtering. Note that after only one deployment, it is not possible to distinguish STFs from LTFs. However, as the number of deployments increases, the number of observations corresponding to STFs in $C'$ approaches zero.

The next step in our algorithm is feature (line) extraction. Line extraction does not rely on any persistent data, using only the filtered composite scan $C'$ to extract a set of lines $L'$. Each $l'_i \in L'$ is defined by endpoints $p_{i_1}$ and $p_{i_2}$, a scatter matrix $S_i$, a center of mass $p_{i_{cm}}$, and a mass $M_i$.

Uncertainties in the endpoints of each line segment are computed by analyzing a distribution of possible endpoints generated via Monte Carlo resampling the initial set of observations and subsequently refitting the samples. For a given line $l_i$ the uncertainty estimation step takes as input the endpoints $p_{i_1}$ and $p_{i_2}$ and a set of inliers $I_i$, and produces covariance estimates $Q_{i_1}$ and $Q_{i_2}$ for these endpoints.

\begin{figure}[!ht]
  \centering
  \includegraphics[scale=0.22, trim = 2mm 0mm 0mm 10mm]{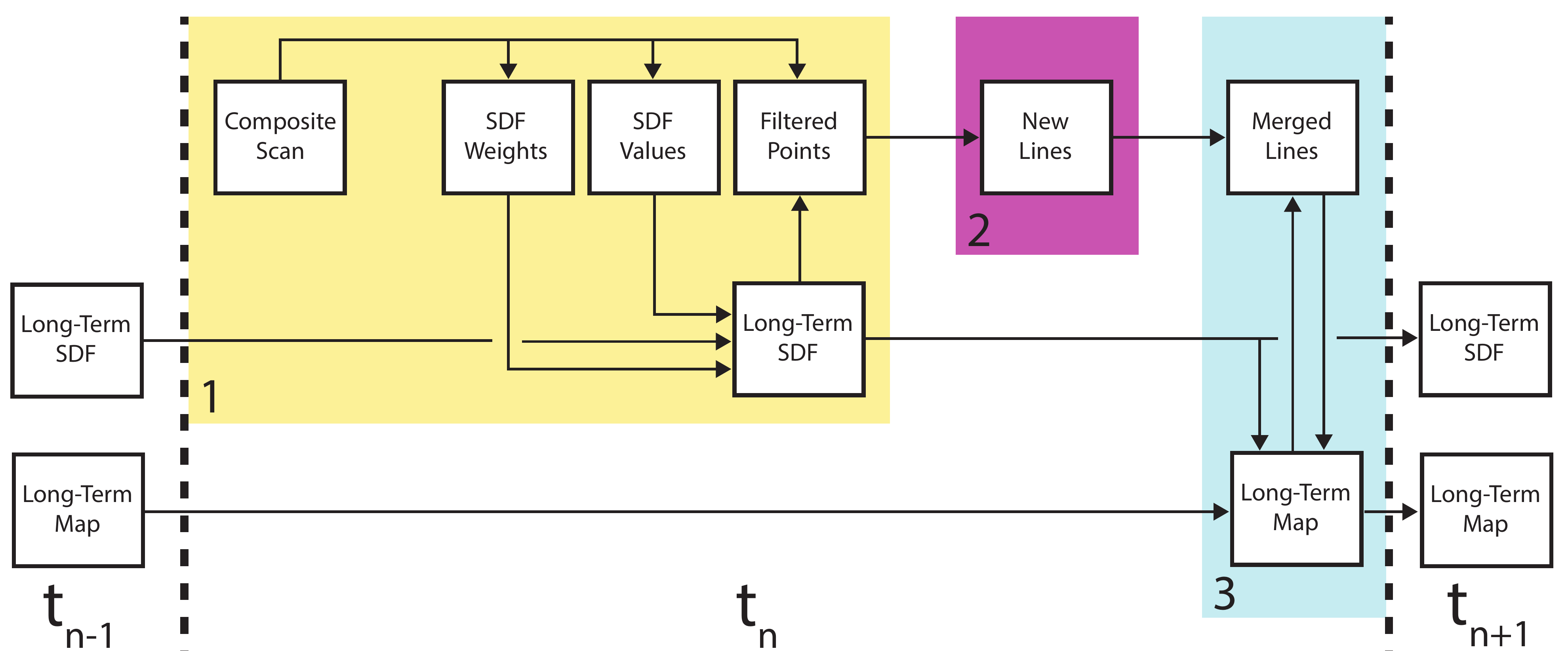}
  \caption{Flow of information during processing of a single deployment, deployment $n$. Boxes 1, 2, and 3 correspond to SDF filtering, line finding and uncertainty estimation, and map updating, respectively.}
  \vspace{-1.2em}
\label{fig:BlockDiag}
\end{figure}

The long-term vector map is updated based on the newest set of extracted lines and the current SDF. Similar lines are merged into a single line, obsolete lines are deleted, and uncertainties are recomputed. Thus, this step takes the results from the most recent deployment, $L'$, as well as the existing map given by $L$, $W^*$, and $V^*$, and outputs an updated map, $L^*$. Fig. \ref{fig:BlockDiag} presents an overview of the algorithm as it operates on data from a single deployment. 

\section{SDF-based Filtering} 

Let $C$ be a composite scan, where $c \in C$ is a single observation providing depth $\rho$ and angle $\alpha$ with respect to the laser's frame and the robot pose $p = (x,y,\theta)$ at which $\rho$ and $\alpha$ were recorded. That is, $c = [\rho, \alpha, p]$. The filtering problem is to determine $C' \subset C$ such that all $c' \in C'$ originate from LTFs and all $c \in C \setminus C'$ originate from STFs and DFs. 

Determining $C'$ is a three-step process. First, we construct a ST-SDF over the area observed by the robot during the deployment corresponding to $C$. Next, we update the LT-SDF based on the ST-SDF. Finally, we use the updated LT-SDF to decide which observations correspond to LTFs.

\subsection{SDF Construction}

SDFs operate over discretized space, so we create a grid of resolution $q$ containing all $c \in C$. Each pixel in the SDF maintains two measurements, a value $d_0$ and a weight $w_0$. For every observation $c \in C$, all pixels that lie along the given laser ray update their respective values according to $d_0 = \frac{w_0 d_0 + wd}{w_0 + w}$ and $w_0 = w_0 + w$, where $d_0$ and $w_0$ are the current distance and weight values, respectively, and $d$ and $w$ are the distance and weight values for the given reading $c$. $d$ and $w$ are given by
 
\begin{figure}[!ht]
  \centering
  \includegraphics[scale = 0.37, trim = 0mm 10mm 0mm 13mm]{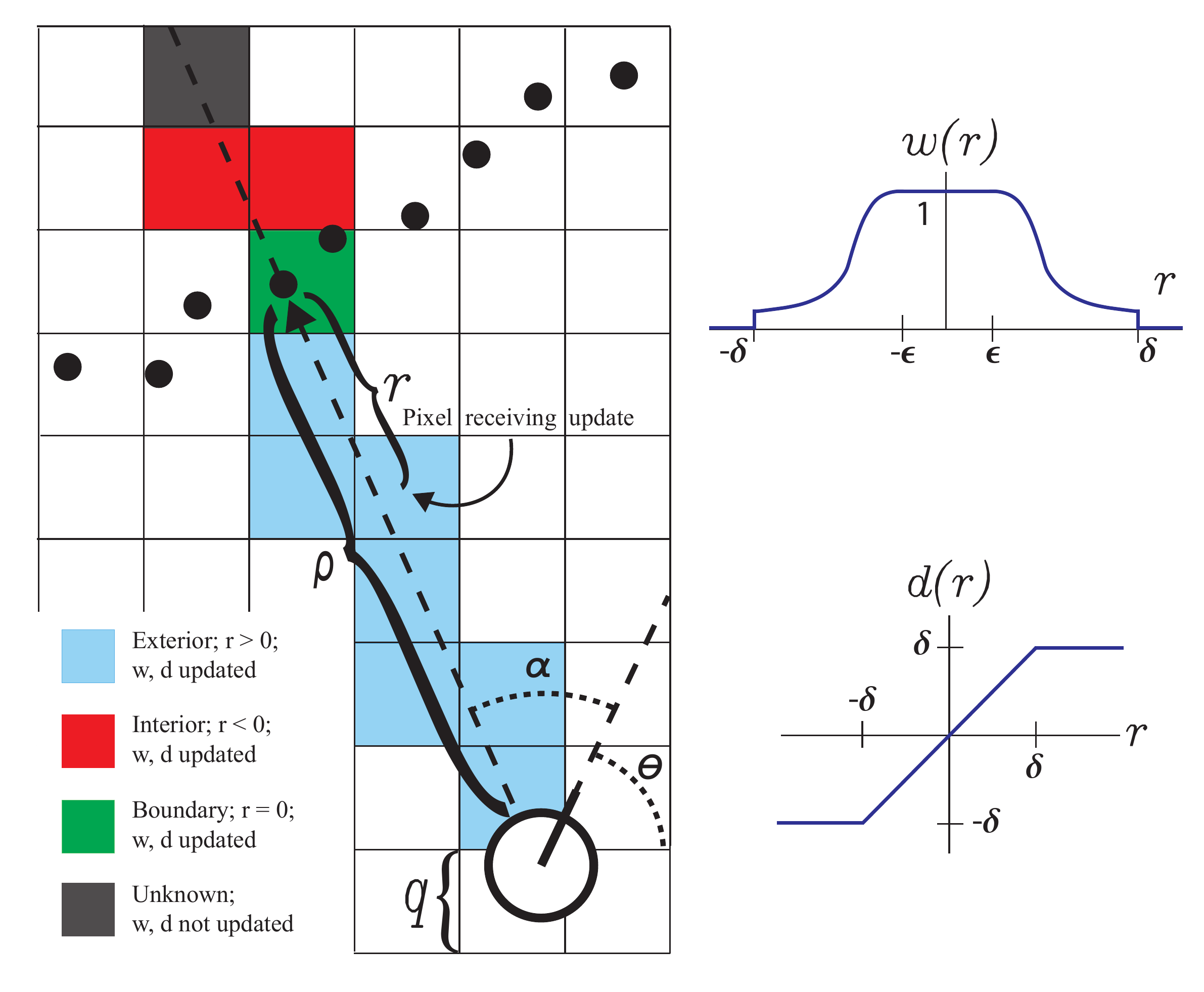}
  \caption{SDF construction from a single laser ray. Pixels along the laser ray are updated if they are free, or if they are just beyond the obstacle. Over many ray casts, pixels may be marked as belonging to more than one category (boundary, interior, exterior) due to sensor noise. The SDF's main advantage is that it ignores erroneous readings.}
\label{fig:SDF}
\end{figure}

\vspace{-5.5mm}
\begin{equation}
d(r) =
  \begin{cases} 
      \delta & \text{if}  \hspace{2mm} r > \delta \\
     r & \text{if} \hspace{2mm} |r| \leq \delta \\
     -\delta & \text{if}  \hspace{2mm} r < -\delta
  \end{cases}, \hspace{2mm}
w(r) = 
  \begin{cases} 
      1 & \text{if}  \hspace{2mm} |r| < \epsilon \\
      e^{G} & \text{if} \hspace{2mm} \epsilon \leq |r| \leq \delta \\
      0 & \text{if}  \hspace{2mm} |r| > \delta,
  \end{cases}
\end{equation}

\noindent where $G = -\sigma (r-\epsilon)^2$ and $r$ is the signed distance from the range reading to the pixel, with pixels beyond the range reading having $r < 0$ and those in front having $r > 0$. $\sigma$ and $\epsilon$ are parameters that depend on the accuracy of the sensor. Pixels that are located along the ray but are more than $\delta$ beyond the detection point are not updated since we do not know whether or not they are occupied. Fig. \ref{fig:SDF} illustrates a single pixel update during SDF construction. Note that this process is parallelizable since weights and values for each pixel may be calculated independently. Thus, the SDF construction step, outlined in Algorithm \ref{alg:sdf}, runs in time proportional to $|C|$.

\vspace{-4mm}

The intuition for the weight and value functions comes from two observations. First, capping $|d(r)|$ by $\delta$ helps keep our SDF robust against anomalous readings. Second, $w(r)$ follows common laser noise models. Other choices for $d(r)$ and $w(r)$ may yield similar results; however, these choices have already been successfully adopted elsewhere  ~\cite{SDFtracking}. 

\subsection{SDF Update}

Once the ST-SDF is calculated we normalize the weights:

\begin{equation}
w_{norm} = 
\begin{cases} 
      0 & \text{if}  \hspace{3mm} \frac{w}{w_{max}} \leq T_1, \\
      1 & \text{otherwise}.
\end{cases}
\end{equation}

\noindent Here, $w_{max}$ is the maximum weight over all pixels and $T_1$ is a dynamic feature threshold. The LT-SDF is then updated as the weighted average of the ST-SDF and the LT-SDF, i.e. $W^* =$\textsc{WeightedAverage}$(W,W')$. Pixel weights loosely map to our confidence about the associated value, and values are an estimate for how far a pixel is from the nearest surface.
\vspace{-4mm}

\subsection{SDF Filter}

Given an up-to-date SDF, we determine $C'$ using bicubic interpolation on the position of each observation $c$ and the updated LT-SDF. The filtering criteria are $c' \in C'$ if \textsc{BicubicInterpolation}$(c,W^*) > T_2$ and \textsc{BicubicInterpolation}$(c,V^*) < T_d$, where $T_2$ is a STF threshold and $T_d$ is a threshold that filters observations believed to be far from the surface of any object. Lines 11-15 in Algorithm \ref{alg:sdf} detail the filtering process. Thus, after running \textsc{SDF Filtering}($C, W, V$), we obtain a filtered composite scan $C'$, used to find LTFs. Additionally, \textsc{SDF Filtering} updates the LT-SDF needed for the map update step.
\vspace{-3mm}

  \begin{algorithm}[ht]
  \caption{\textsc{SDF Filtering}}
  \begin{algorithmic}[1]
  \State $\textbf{Input:}$ Raw composite scan $C$, long-term SDF weights $W$ and values $V$
  \State $\textbf{Output:}$ Filtered composite scan $C'$, updated SDF weights $W^*$ and values $V^*$
  \State $V' \gets$ empty image
  \State $W' \gets$ empty image
  \ForAll{range readings $c \in C$} 
  \State $V' \gets$ \textsc{valueUpdate}$(W',c)$
  \State $W' \gets$ \textsc{weightUpdate}$(W',c)$ 
  \EndFor
  \State $W' \gets$ \textsc{normalize}$(W')$
  \State $W^*, V^* \gets \textsc{updateSDF}(V',W')$
  \State $C' \gets \emptyset$
  \ForAll{range readings $c \in C$}
  \State $b_w \gets$ \textsc{BicubicInterpolation}$(W^*, c)$  
  \State $b_v \gets$ \textsc{BicubicInterpolation}$(V^*, c)$
  \If{$b_w > T_w$ \text{\bf{and}} $b_v < T_v$}
  \State $C' \gets C' \cup c$
  \EndIf
  \EndFor
  \end{algorithmic}
  \label{alg:sdf}
  \end{algorithm}

\vspace{-6mm}

\section{Line Extraction}

Given $C'$, extracting lines $l_1 \ldots l_n$ requires solving two problems. First, for each $l_i$, a set of observations $C_i \subset C'$ must belong to line $l_i$. Second, endpoints $p_{i_1}$ and $p_{i_2}$ defining line $l_i$ must be found. We take a greedy approach, utilizing sequential local RANSAC to provide plausible initial estimates for line endpoints $p_1$ and $p_2$. Points whose distance to the line segment $\overline{p_1 p_2}$ is less than $T_r$, where $T_r$ is proportional to the noise of the laser, are considered members of the inlier set $I$. Once a hypothetical model and set of inliers with center of mass $p_{cm}$ have been suggested by RANSAC (lines 5-7 in Algorithm \ref{alg:ransac}), we perform a non-linear least-squares optimization using the cost function
\vspace{-3mm}

\begin{equation}
\begin{aligned}
R = & \frac{||p_{cm} - p_1|| + ||p_{cm} - p_2||}{|I|} \\
+ &
 \begin{cases} 
      ||p-p_2|| & \text{if}  \hspace{3mm} t < 0 \\
      ||p-p_1|| & \text{if}  \hspace{3mm} t > 1 \\
      ||p'_1 + t(p'_2 - p'_1) - p|| & \text{otherwise}
  \end{cases}.
\end{aligned}
\end{equation}

\noindent The new endpoints $p'_1$ and $p'_2$ are then used to find a new set of inliers $I'$. $t= \frac{(p - p'_1) \cdot (p'_2 - p'_1) }{||p'_2 - p'_1||^2}$ is the projection of a point $p \in I$ onto the infinite line containing $p'_1$ and $p'_2$. Iteration terminates when $||p_1 - p'_1|| + ||p_2 - p'_2|| < T_c$, where $T_c$ is a convergence threshold. 

This cost function has several desireable properties. First, when all points lie between the two endpoints, the orientation of the line will be identical to the least-squares solution. Second, when many points lie beyond the ends of the line segment, the endpoints are pulled outward, allowing the line to grow and the region in which we accept inliers to expand. Last, the $\frac{||p_{cm} - p_1|| + ||p_{cm} - p_2||}{|I|}$ term allows the line to shrink in the event that the non-linear least-squares solver overshoots the appropriate endpoint. Once a set of lines $L'$ has been determined by running \textsc{Line Extraction}($C'$) we complete our analysis of a single deployment by estimating our uncertainty in feature locations.
\vspace{-3mm}

  \begin{algorithm}[ht]
  \caption{\textsc{Line Extraction}}
  \begin{algorithmic}[1]
  \State $\textbf{Data:}$ Filtered composite scan $C'$
  \State $\textbf{Result:}$ Set of new lines $L'$
  \State $L' \gets \emptyset$
  \While{$C'$ {\bf not} empty}
  \State Propose $p_1, p_2$ via RANSAC
  \State $I \gets$ \textsc{findInliers}($p_1$, $p_2$)
  \State $p'_1$, $p'_2 \gets$ \textsc{fitSegment}($I$)
  \While {$||p'_1 - p_1|| + ||p'_2 - p_2|| > T_C$}
  \State $I \gets$ \textsc{findInliers}($p'_1$, $p'_2$)
  \State $p_1, p_2 \gets p'_1, p'_2$
  \State $p'_1$, $p'_2 \gets$ \textsc{fitSegment}($I$)
  \EndWhile
  \State $L' \gets L' \cup \overline{p'_1 p'_2}$
  \State $C' \gets C' \setminus I$
  \EndWhile
  \end{algorithmic}
  \label{alg:ransac}
  \end{algorithm}
\vspace{-4mm}

\section{Uncertainty Estimation}

\begin{figure}[!ht]
  \centering
  \includegraphics[scale=0.4, trim=0mm 0mm 0mm 14mm]{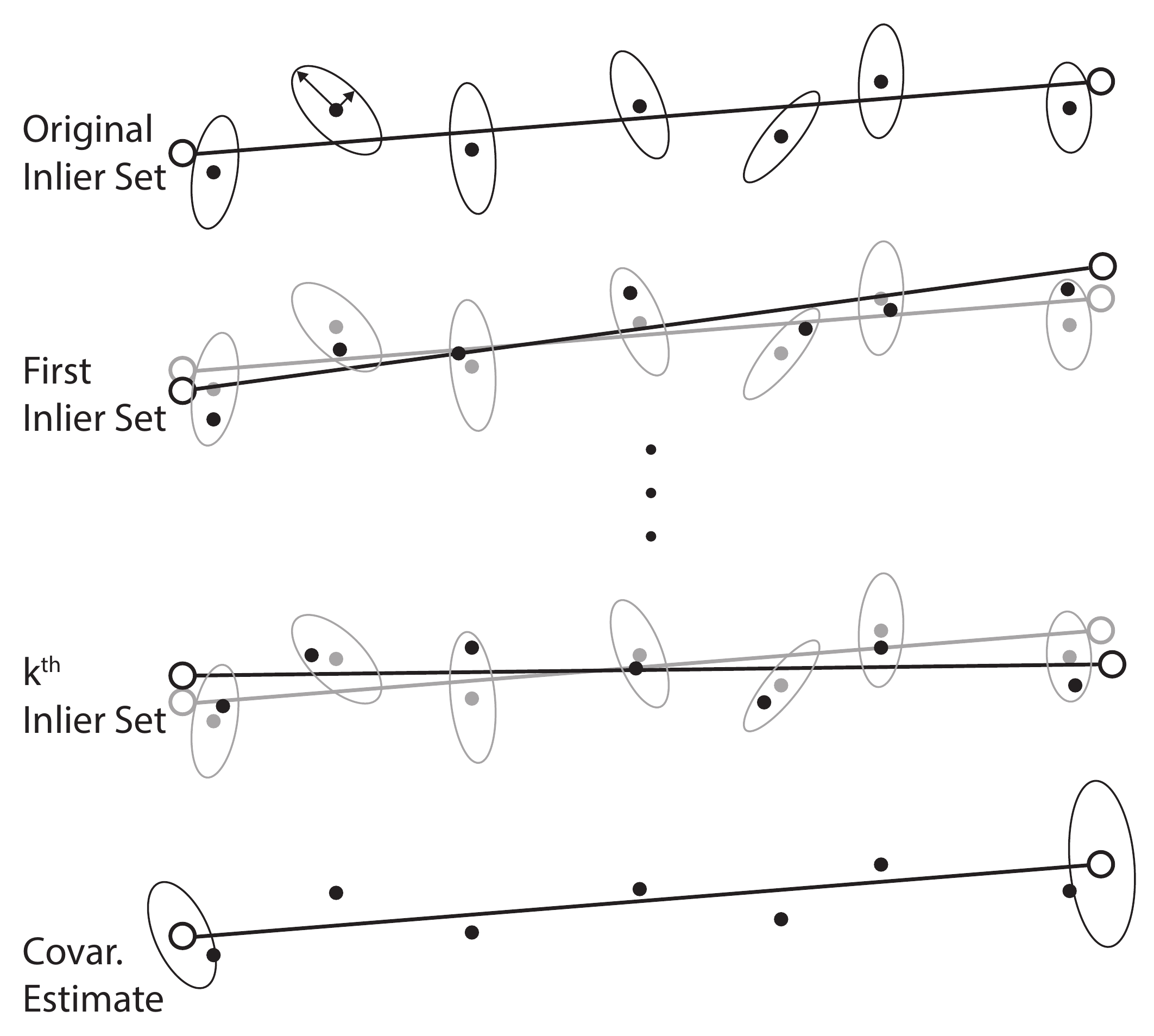}
  \caption{Monte Carlo uncertainty estimation of feature endpoints. Given an initial set of observations and their corresponding covariances represented by ellipses, we resample the observations and fit a line $k$ times. The resulting distribution of endpoints is used to estimate endpoint covariance. \vspace{-8mm}}
\label{fig:MonteCarlo}
\end{figure}

Given a line $l_i$, with endpoints $p_{i_1}$ and $p_{i_2}$ and a set of inliers $I_i$, uncertainty estimation produces covariance estimates $Q_{i_1}$ and $Q_{i_2}$ for $p_{i_1}$ and $p_{i_2}$, respectively. To estimate $Q_{i_1}$ and $Q_{i_2}$ we resample $c_i \sim I_i$ using the covariance $Q_i ^c$ of each range reading. $Q_i ^c$ is derived based on the sensor noise model in ~\cite{noise}. In world coordinates $Q_i ^c$ is given by

\begin{equation}
\begin{aligned}
Q_i ^c = & \frac{\rho^2 \sigma_{\alpha}^2}{2} 
\left[ \begin{array}{cc}
2 \text{sin}^2 (\alpha + \theta) & - \text{sin} (2(\alpha + \theta)) \\
- \text{sin} (2(\alpha + \theta)) & 2 \text{cos}^2 (\alpha + \theta) \end{array} \right] \\
+ & \frac{\sigma_{\rho}^2}{2}
\left[ \begin{array}{cc}
2 \text{cos}^2 (\alpha + \theta) & \text{sin} (2(\alpha + \theta))  \\
 \text{sin} (2(\alpha + \theta)) & 2 \text{sin}^2 (\alpha + \theta)  \end{array} \right],
\end{aligned}
\end{equation}

\noindent where $\sigma_{\rho}$ and $\sigma_{\alpha}$ are standard deviations for range and angle measurements for the sensor, respectively. Resampling $k$ times, as shown in Fig. \ref{fig:MonteCarlo}, produces a distribution of $p_1$ and $p_2$. We then construct a scatter matrix $S$ from the set of hypothetical endpoints, and compute covariances $Q_1$ and $Q_2$ by using the SVD of $S$.

The Monte Carlo approach, detailed in Algorithm \ref{alg:montecarlo}, is motivated by the following factors: 1) There is no closed-form solution to covariance for endpoints of a line segment. 2) A piece-wise cost function makes it difficult to calculate the Jacobian reliably. 3) Resampling is easy since we already have $I_i$ and can calculate $Q_i ^c$.

  \setlength{\textfloatsep}{0pt}
  \begin{algorithm}[ht]
  \caption{\textsc{Feature Uncertainty Estimation}}
  \begin{algorithmic}[1]
  \State $\textbf{Input:}$ Set of new lines $L'$, number of samples $k$
  \State $\textbf{Output:}$ Set of endpoint covariance estiamates $Q'_1$, $Q'_2$
  \State $Q'_1 \gets \emptyset$, $Q'_2 \gets \emptyset$
  \ForAll{$l'_i \in L'$}
  \State $P'_1 \gets \emptyset$, $P'_2 \gets \emptyset$
  \State $I_i \gets$ inliers associated with $l'_i$
  \For{$k$ iterations}
  \State $I'_i \gets \emptyset$
  \ForAll{$c \in I_i$}
  \State $c' \gets$ \textsc{sample}$(I_i, c, Q_i ^c)$
  \State $I'_i \gets I'_i \cup c'$
  \EndFor
  \State $p'_1$, $p'_2 \gets$ \textsc{fitSegment}($I'_i$)
  \State $P'_1 \gets P'_1 \cup p'_1$, $P'_2 \gets P'_2 \cup p'_2$
  \EndFor
  \State $Q'_1 \gets Q'_1 \cup$ \textsc{estimateCovariance}$(P'_1)$
  \State $Q'_2 \gets Q'_2 \cup$ \textsc{estimateCovariance}$(P'_2)$
  \EndFor
  \end{algorithmic}
  \label{alg:montecarlo}
  \end{algorithm}

\section{Map update}
\vspace{-1mm}

Given the current map $L$, the LT-SDF, and a set of new lines, $L'$, where every $l_i \in L'$ is specified by a set of endpoints $p_{i_1}$ and $p_{i_2}$, a set of covariance matrices $Q_{i_1}$ and $Q_{i_2}$, a center of mass $p_{i_{cm}}$, a mass $M_i$, and a partial scatter matrix $S_i$, the update step produces an updated map $L^*$. 

The map updates are divided into two steps outlined in Algorithm \ref{alg:update}. First, we check if all current lines in the map are contained within high-weight regions. That is, we check that the weight $w_{xy}$ of every pixel a given line passes through satisfies $w_{xy} \geq T_2$. If a line lies entirely within a high-weight region, it remains unchanged. Similarly, if a line lies entirely outside all high-weight regions, it is removed. If only part of a line remains within a high-weight region, we can lower bound the mass of the remaining region by $M' = M \frac{||p'_1 - p'_2||}{||p_1 - p_2||}$, where $p'_1$ and $p'_2$ are the extreme points of the segment remaining in the high-weight region (line 11). We then resample $M'$ points uniformly along the line between $p'_1$ and $p'_2$, adding Gaussian noise in the perpendicular direction with a standard deviation $\sigma$ based on the sensor noise model. The points are then fit using the cost function given in (4). The sampling and fitting process is executed a fixed number of times (lines 12-16), and the distribution of fit results is then used to compute covariance estimates for the new line. 

The second part of the update involves merging new lines, $L'$, with lines from the current map, $L$. This process consists of first finding candidates for merging (lines 22-23) and then computing the updated parameters and covariances (lines 24-25). Note that the mapping from new lines to existing lines may be onto, but without loss of generality we consider merging lines pairwise. Because our parameterization uses endpoints, lines which ought to be merged may not have endpoints near one another. So, we project $p'_{i_1}$ and $p'_{i_2}$ from $l'_i$ onto $l_j$, resulting in $p_{j_1}^{proj}$ and $p_{j_2}^{proj}$, respectively. $l'_i$ and $l_j$ are merged if they pass the chi-squared test:
\begin{equation}
\chi^2 = \Delta l_k^T (Q_{j_k} ^{int} + Q'_{i_k}) \Delta l_k < T_{\chi}, \hspace{5mm} k = 1,2
\end{equation}

\noindent where $\Delta l_k = p'_{i_k} - p_{j_k}^{proj}$, and $Q_{j_k}^{int}$ is given by a linear interpolation of the covariance estimates for $p_{j_1}$ and $p_{j_2}$ determined by where $p'_{i_k}$ is projected along $l_j$. 

We would like our merged LTFs and their uncertainty measures to remain faithful to the entire observation history. However, storing every observation is infeasible. Instead, our method implicitly stores observation histories using decoupled scatter matrices ~\cite{scatter}, reducing the process of merging lines with potentially millions of supporting observations to a single matrix addition. 

The orientation of the new line is found via eigenvalue decomposition of the associated scatter matrix, and new endpoints are found by projecting the endpoints from the original lines onto the new line and taking the extrema. Thus, after executing \textsc{Map Update}($L',L,W^*,V^*$), we have a set of vectors $L^*$ in $\mathbb{R}^2$ corresponding to LTFs.

Table \ref{tab:gt} displays the major parameters and physical constants needed for long-term vector mapping.

\begin{figure*}[t!]
	\centering
	\vspace{-1.0em}
	\subfloat[]{\label{fig:scene5_scene}
	\includegraphics[height=2.7cm]{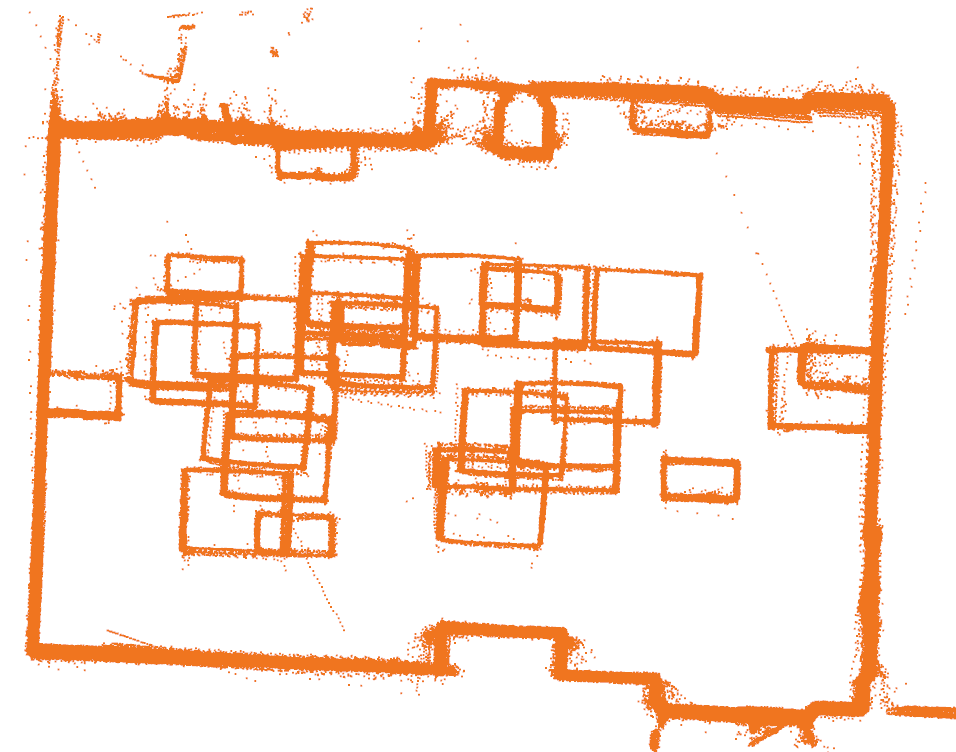}}
      \subfloat[]{\label{fig:scene5_sc}
      \includegraphics[height=2.7cm]{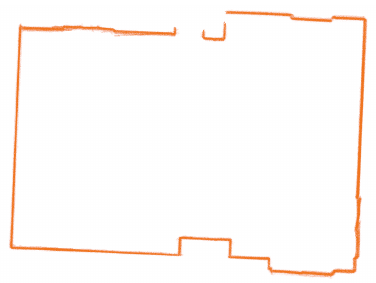}}
      \subfloat[]{\label{fig:scene5_}
      \includegraphics[height=2.7cm]{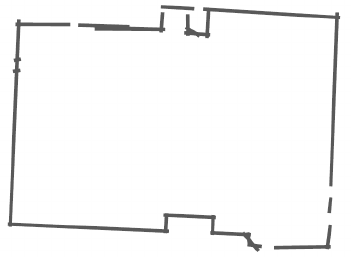}}
	\vspace{-0.5em}
      \subfloat[]{\label{fig:scene3_scene}
	\includegraphics[height=3.0cm]{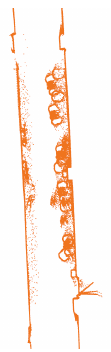}}
      \subfloat[]{\label{fig:scene3_sc}
      \includegraphics[height=3.0cm]{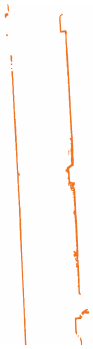}}	
	\subfloat[]{\label{fig:scene3_}
      \includegraphics[height=3.0cm]{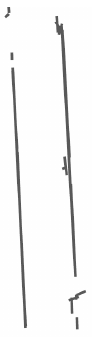}}
	\vspace{-0.5em}
	\subfloat[]{\label{fig:scene6_scene}
	\includegraphics[height=2.7cm]{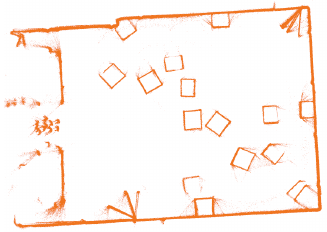}}
      \subfloat[]{\label{fig:scene6_sc}
      \includegraphics[height=2.7cm]{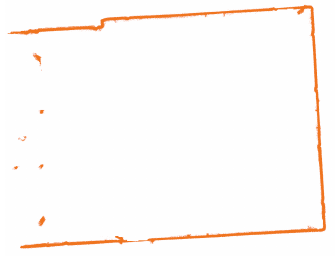}}	
      \subfloat[]{\label{fig:scene6_}
      \includegraphics[height=2.7cm]{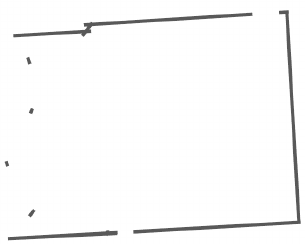}}
	\vspace{-0.5em}
      \subfloat[]{\label{fig:scene4_scene}
	\includegraphics[height=2.7cm]{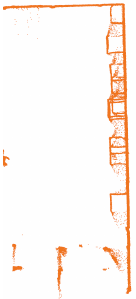}}
      \subfloat[]{\label{fig:scene4_sc}
      \includegraphics[height=2.7cm]{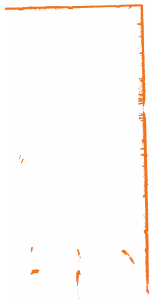}}	
	\subfloat[]{\label{fig:scene4_}
      \includegraphics[height=2.7cm]{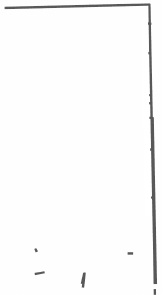}}

	\caption{Raw data, filtered data, and resultant maps for MIT (a-c), AMRL (d-f), Wall-occlusion (g-i) and Hallway-occlusion (j-l) datasets. Data shown in the left and center columns are aggregates of all deployments and are never actually stored while the algorithm is operating. The last column is the resultant LTVM which is stored in full, requiring only a few KB.  Note the absence of STFs from the final maps, as well as the presence of doorways. In the MIT dataset, some doors were open only once over the entire deployment history. The Hallway-occlusion dataset demonstrates the algorithm's robustness to STFs, as it is able to distinguish the column in the hallway even as it is partially or completely occluded on every deployment.}
  \label{fig:scenes}
	\vspace{-7mm}
\end{figure*}

\vspace{-3mm}

\begin{table}[h]
\caption{Thresholds and physical constants}
\centering
\begin{tabular}{| l | l | l | l |}
\hline
Name & Symbol & Domain & Our Value \\
\hline
DF Threshold & $T_1$ & $(0,1)$ & $0.2$ \\
STF Threshold & $T_2$ & $(0,1)$ & $0.95$ \\
Line Merge Criteria & $T_{\chi^2}$ & $>0$ & $30$ \\
Sensor Noise Threshold & $T_d$ & $>0$ & $0.05$ meters \\
RANSAC Inlier Criteria & $T_r$ & $>0$ & $0.12$ meters \\
Line Fit Convergence & $T_c$ & $>0$ & $0.05$ meters \\
SDF Max Value & $\delta$ & $>0$ & $0.2$ meters \\
\hline
\end{tabular}
\label{tab:gt}
\end{table}

\begin{algorithm}[ht]
  \caption{\textsc{Map Update}}
  \begin{algorithmic}[1]
  \State $\textbf{Input:}$ Set of new lines $L'$, set of long-term lines $L$, long-term SDF $W$, $V$
  \State $\textbf{Output:}$ Updated long-term lines $L^*$
  \ForAll{$l_i \in L$}
  \State trace along $l_i$
  \If{$l_i$ exists entirely outside high-weight regions}
  \State $L \gets L \setminus l_i$
  \EndIf
  \If{$l_i$ exists partially outside high-weight regions}
  \State $L \gets L \setminus l_i$
  \State $P'_1, P'_2 \gets \emptyset$
  \ForAll{remaining parts of $l_i$, $\partial l_i$}
  \State $p_1, p_2 \gets$ \textsc{getEnpoints}($\partial l_i$)
  \For{$k$ iterations}
  \State $I \gets$ \textsc{regenerateInliers}($p_1, p_2$)
  \State $p'_1, p'_2 \gets$ \textsc{fitSegment}($I$)
  \State $P'_1 \gets P'_1 \cup p'_1$
  \State $P'_2 \gets P'_2 \cup p'_2$
  \EndFor
  \State \textsc{estimateCovariance}($P'_1, P'_2$)
  \State $L \gets L \cup \partial l_i$
  \EndFor
  \EndIf
  \EndFor
  \State $L^* \gets \emptyset$
  \ForAll{$l'_i \in L'$}
  \ForAll{$l_j \in L$}
  \State $\chi^2 \gets \Delta l_k^T (Q_{j_k} ^{int} + Q'_{i_k}) \Delta l_k < T_{\chi}, \hspace{5mm} k = 1,2$
  \If{$\chi^2 < T_{\chi ^2}$ for $k=1, 2$}
  \State $l_j \gets$ \textsc{merge}$(l_j, l'_i)$
  \State $L^* \gets L^* \cup l_j$
  \Else
  \State $L^* \gets L^* \cup l'_i$
  \EndIf
  \EndFor
  \EndFor
  \end{algorithmic}
  \label{alg:update}
\end{algorithm}

\vspace{-5mm}

\section{Results}

To demonstrate the effectiveness of our algorithm we mapped 4 different environments, including standard data. Data was also collected from three separate environments, in addition to the MIT Reading Room: AMRL, Wall-occlusion, and Hallway-occlusion, using a mobile robot platform and a Hokuyo UST-10LX laser range finder. The AMRL, Wall, and Hall datasets consist of 8, 5, and 5 deployments, respectively. MIT Reading Room contains 20 deployments. Each deployment contains hundreds of scans, corresponding to hundreds of thousands of observations. Datasets are intended to display a high amount of clutter and variability, typical of scenes mobile robots need to contend with. 

The MIT and AMRL data sets are intended to test the accuracy of our algorithm over larger numbers of deployments. Both rooms contain multiple doors, walls of varying lengths, and achieve many different configurations, testing our ability to accurately identify LTF positions. The Wall- and Hallway-occlusion datasets are intended to measure our algorithm's robustness in environments where LTFs are heavily occluded. 

Quantitatively we are concerned with accuracy and robustness, defining accuracy as pairwise feature agreement, where inter-feature distances are preserved, and feature-environment correspondance, where vectors in the map correspond to LTFs in the environment. Vectors should also be absent from areas where there are no long-term features such as doorways. Metric ground truth is established by either measuring wall lengths or hand-fitting the parts of the data we know correspond to LTFs. Robustness refers to a map's ability to deal with large numbers of STFs and lack of degradation over time.  

Over all datasets, our approach correctly identifies all 7 doorways (4 in MIT, 2 in AMRL, 1 in Hallway), and correctly ignores all 73 STFs. Using the AMRL and MIT datasets, we find the average difference between pair-wise feature separation in the generated map versus ground truth to be on the order of 2cm. Our line extraction method yields MSE values in the order of $0.0001$ meters, while marching squares yields a MSE of roughly $0.0003$, about three times as large. Additionally, marching squares yields over $3000$ features while LTVM maintains on the order of $30$ LTFs. Furthermore, the maps do not degrade over the timescales tested in this paper, with no noticeable difference in average pair-wise feature disagreement between maps generated after one deployment and those considering all deployments. Fig. \ref{fig:scenes} displays qualitative results for all environments.

\vspace{-2mm}
\section{Conclusion}

In this paper, we introduced an SDF-based approach to filter laser scans over multiple deployments in order to build and maintain a long-term vector map. We experimentally showed the accuracy and robustness of the generated maps in a variety of different environments and further evaluated the effectiveness of long-term vector mapping compared to more standard, occupancy grid techniques. Currently this approach processes observations in batch post deployment, but modifying the algorithm to run in an online setting seems an appropriate next step. 

\vspace{-2mm}
\section{Acknowledgements}

The authors would like to thank Manuela Veloso from Carnegie Mellon University for providing the CoBot4 robot used to collect the AMRL datasets, and to perform experiments at University of Massachusetts Amherst.


\bibliography{SDFVectorMapping}

\end{document}